\documentclass{article}
\usepackage{ijcai21}
\usepackage{xcolor}
\usepackage{lipsum}
\usepackage{stfloats}
\usepackage{times}
\usepackage{soul}
\usepackage{url}
\usepackage[hidelinks]{hyperref}
\usepackage[utf8]{inputenc}
\usepackage[small]{caption}
\usepackage{graphicx}
\usepackage{amsmath}
\usepackage{multirow}
\usepackage{amsthm}
\usepackage{booktabs}
\usepackage{algorithm}
\usepackage{algpseudocode}

\newtheorem{remark}{Remark}

\title{A Clustering-based Framework for Classifying Data Streams}

\author{
Xuyang Yan$^1$\and
Abdollah Homaifar$^1$\thanks{Contact Author}\and
Mrinmoy Sarkar$^1$\and \\
Abenezer Girma$^{1}$\And
Edward Tunstel$^2$
\affiliations
$^1$North Carolina A\&T State University, Greensboro, NC, 27401, USA\\
$^2$Raytheon Technologies Research Center, East Hartford, CT, 06108, USA
\emails
xyan@aggies.ncat.edu,
homaifar@ncat.edu,
msarkar@aggies.ncat.edu,\\
aggirma@aggies.ncat.edu,
tunstel@ieee.org
}

\begin{document}

\maketitle

\begin{abstract}
  The non-stationary nature of data streams strongly challenges traditional machine learning techniques. Although some solutions have been proposed to extend traditional machine learning techniques for handling data streams, these approaches either require an initial label set or rely on specialized design parameters. The overlap among classes and the labeling of data streams constitute other major challenges for classifying data streams. In this paper, we proposed a clustering-based data stream classification framework to handle non-stationary data streams without utilizing an initial label set. A density-based stream clustering procedure is used to capture novel concepts with a dynamic threshold and an effective active label querying strategy is introduced to continuously learn the new concepts from the data streams. The sub-cluster structure of each cluster is explored to handle the overlap among classes. Experimental results and quantitative comparison studies reveal that the proposed method provides statistically better or comparable performance than the existing methods.
\end{abstract}
\section{Introduction}
The recent advances in information technology have led to an increasing trend in the application of data streams, and data stream classification has captured the attention of many researchers. Unlike traditional classification problems, data streams are continuously evolving such that the data distributions can change dynamically and new data patterns may appear. This non-stationary nature of data streams, known as \textit{concept drift} and \textit{concept evolution} \cite{masud2010classification,gama2014survey}, requires a continuous learning capability for traditional machine learning techniques to handle data streams \cite{aggarwal2006framework,bouchachia2011incremental,mu2017streaming}. The scarcity of data labels and expensive labeling costs also limit the conventional classification techniques \cite{din2020online}. Besides, limited time and memory space pose an additional layer of challenges for classifying the large volume of data streams. \\
\indent To address these challenges, a great number of studies have been conducted on classifying data streams considering two primary aspects, data stream classification through: (i) supervised and (ii) semi-supervised learning. For supervised data stream classification frameworks, most studies \cite{bifet2009new,bifet2009adaptive,bifet2010leveraging,losing2018tackling,gomes2017survey} focus on extending traditional offline machine learning techniques with an assumption that the label of each incoming sample will be available after it is classified. However, this assumption does not always hold in real-world data stream classification problems and the labeling of data streams is time-consuming and expensive. Additionally, to adapt to the change in data streams, the optimization of the learnable parameters also imposes high computational overhead for supervised approaches. \\
\indent Semi-supervised approaches \cite{masud2012facing,shao2016reliable,wagner2018semi,zhu2020semi,din2020online} are developed as an alternative solution to perform  data stream classifications using a small portion of labeled data and have shown substantial success. However, several challenges still remain. First, the detection of a novel class strongly depends on certain threshold parameters and the determination of threshold parameters is an open challenge. Secondly, the overlap among different classes is a common problem in data stream classification problems while very little research work has been conducted on handling data streams with overlapped classes. Thirdly, most of the existing semi-supervised classification frameworks classify data streams with partially labeled data in a passive learning manner. For better classification performance, this requires an effective labeling strategy to actively guide the learning procedure.\\
\indent Considering the limitations of the existing methods described earlier, we propose a \textbf{C}lustering-based \textbf{D}ata \textbf{S}tream \textbf{C}lassification framework with \textbf{A}ctive \textbf{L}earning, namely CDSC-AL, to handle non-stationary data streams. The proposed framework consists of three main steps: (i) concept drift and novel class detection through clustering, (ii) novel concept learning and concept drift adaptation through active learning, and (iii) classification of data streams using label propagation. With these three steps, CDSC-AL aims to perform non-stationary data stream classification by reducing the labeling costs and handling the overlap among classes. In summary, the contributions of this paper are as follows:
\begin{itemize}
    \item Extend a density-based data stream clustering method to capture \textit{concept drift} and \textit{concept evolution} for non-stationary data streams with a dynamic threshold.
    \item Develop an effective distance-based active learning procedure to query a small portion of labels for adapting to changes in data streams.
    \item Propose a classification procedure to handle the overlap among classes by investigating the sub-cluster structure inside clusters. 
\end{itemize}

The remainder of this paper is organized as follows. Section \ref{sec:2} provides a review of the related work. The details of the CDSC-AL framework are discussed in Section \ref{sec:3} and the complexity analysis of the CDSC-AL framework is described in Section \ref{sec:4}. Section \ref{sec:5} presents the experimental results and comparison studies with the state-of-the-art methods. Concluding remarks and future work are outlined in Section \ref{sec:6}. 
\section{Related Work} \label{sec:2}
Over the past few decades, great efforts have been conducted to adapt offline supervised learning techniques for classifying streaming data. In \cite{aggarwal2006framework}, the authors proposed a dynamic data classification framework using traditional classifiers to classify streaming data. A general framework was developed for adapting conventional classification techniques to perform data stream classification using clustering in \cite{masud2010classification,masud2012facing}. In \cite{bifet2007learning,bifet2013efficient}, ADaptive sliding WINdow (ADWIN) is proposed as a detector to capture the change of data streams and work collaboratively with traditional classification techniques for handling data streams. Later, ADWIN was widely used in ensemble-based data stream classification frameworks \cite{bifet2010leveraging,gomes2017survey}. To enhance the time and memory efficiency, a Self Adjusting Memory (SAM) model is introduced for offline classifiers to deal with non-stationary data streams in \cite{losing2018tackling}. 

Unlike supervised approaches, semi-supervised methods utilize a more realistic assumption on the availability of labels for data streams. The Linear Neighborhood Propagation \cite{wang2007label} method is used to perform data stream classification using a limited amount of labeled data. In \cite{masud2012facing}, ECSMiner is proposed to employ the clustering procedure for data stream classification in a semi-supervised manner. A combination of semi-supervised Support Vector Machines and k-means clustering is employed in \cite{zhang2010classifier} for classifying data streams. In \cite{hosseini2016ensemble}, data streams are divided into chunks and an ensemble of cluster-based classifiers is used to propagate labels among each chunk. The Co-forest algorithm \cite{li2007improve} is extended for data stream classification using a small amount of labeled data in \cite{wang2018improving}.

Recently, online graph-based semi-supervised data stream classification frameworks were investigated in \cite{ravi2016large,wagner2018semi}. In \cite{ravi2016large}, the authors continuously maintained a graph using both the labeled and unlabeled data over time and then predicted the label for incoming unlabeled instances through label propagation. Another novel online graph-based semi-supervised approach is introduced to reduce the time and memory complexity \cite{wagner2018semi}. Primarily, it utilizes a temporal label propagation procedure to not only label the new instances but also to learn from them. 

Few recent studies on data stream classification incorporate active learning into the semi-supervised learning framework \cite{lughofer2016recognizing,mohamad2018active}. Such approaches handle the high labeling costs of data streams by actively querying labels for a small subset of interesting samples. In our proposed framework, we introduce a new active learning strategy to address the high labeling cost issue and combine it with a label propagation procedure to classify data streams in a semi-supervised manner. 

\section{Proposed Approach} \label{sec:3}
In this section, the basic notations and assumptions for the CDSC-AL framework are introduced first. Then, we provide an overview of the CDSC-AL framework and discuss its three main components: concept drift and evolution detection through clustering, the adaptation of drifted and novel concepts using active learning, and classification through label propagation. 
\subsection{Notations and Assumptions}
\paragraph{Notations.}Let $DS$ be an unknown data stream and $CH_{t}$ is a chunk of data from $DS$ such that $DS = \bigcup^{ \infty}_{t=1}\{CH_{t}\}$ where $t$ refers to the time index of a data chunk. For each data chunk, $CH_{t} = \bigcup^{|CH_{t}|}_{i=1}\{\mathbf{x^{i}_{t}}|\mathbf{x^{i}_{t}}\in \mathcal{R}^{m}\}$ where $\mathbf{x^{i}_{t}}$ denotes a data sample in $CH_{t}$ and $m$ is the dimension of $\mathbf{x^{i}_{t}}$. Assume $HS_{t}=\{GS_{t},LS_{t}\}$ is the summary of a data stream where $GS_{t}$ and $LS_{t}$ represent the macro-level and the micro-level summary of $DS$. We use $Z^{i}_{t}$ to represent the $i^{th}$ cluster center of $DS$ at time $t$, and $SC^{i,j}_{t}$ refers to the $j^{th}$ sub-cluster center of cluster $i$ at time $t$. The notation $y^{i,j}_{t}$ denotes the class label of $SC^{i,j}_{t}$.
 
\paragraph{Assumptions.}Considering the non-stationary nature of data streams, the proposed CDSC-AL framework assumes:
\begin{itemize}
    \item \textcolor{black}{The characteristics of data streams can change abruptly or gradually}.
    \item Multiple novel concepts may arise simultaneously.
    \item Overlapped classes will appear over time.
\end{itemize}

\subsection{An Overview of CDSC-AL Framework}
A general algorithm description of the CDSC-AL framework is presented in Algorithm \ref{algo:1}. We extended the recently developed density-based stream clustering algorithm, namely dynamic fitness proportionate sharing (DFPS-clustering) \cite{yan2019novel,yan2020efficient}, to perform the classification of data streams considering several aspects. First, a new merge procedure between clusters from the incoming data chunk and historical clusters is employed and the modified merge procedure is used to detect novel classes and drifted classes. Second, two levels of cluster summary are maintained continuously to reflect the characteristics of the data stream through an active learning procedure. Third, an effective classification procedure using the k-nearest-neighbor (KNN) rule \cite{altman1992introduction} is introduced to classify the incoming data chunk based on the summary and queried labels. Additionally, the overlap among classes is addressed by exploring the sub-cluster information from the micro-level summary.
\begin{algorithm}[tb]
\footnotesize
\caption{An overview of CDSC-AL framework}
\label{algo:1}
\textbf{Input}: $DS$\\
\textbf{Parameters}: $C_{t}$: the set of clusters at time $t$; $CH_{t}$: the current data chunk; $C_{CH_{t}}$: a set of clusters discovered in $CH_{t}$; $Q_{t}$: a small set of samples for active label querying in $CH_{t}$; $Y_{Q_{t}}$: the label set of the queried samples $Q_{t}$; $Y_{CH_{t}}$: the label set for $CH_{t}$.\\
\textbf{Output}: $HS_{t}$
\begin{algorithmic}[1]
\For{$t=1$ to $\infty$}
\State Conduct recursive density evaluation on $CH_{t}$ and rank all samples of $CH_{t}$ according to their density values
\State Perform the search of possible clusters in $CH_{t}$  
\State Merge highly overlapped clusters to obtain $C_{CH_{t}}$ 
\If{$t==1$}
\State $HS_{t}=\emptyset$, $C_{t}=C_{CH_{t}}$
\State $\left[Q_{t},Y_{Q_{t}}\right]$ =  \textsc{ActiveQuery}$(HS_{t},C_{t},CH_{t})$
\State $Y_{CH_{t}}$=\textsc{Classify}$(HS_{t},Q_{t},Y_{Q_{t}},CH_{t})$
\State $HS_{t}$=\textsc{ClusteringModel}$(HS_{t},C_{CH_{t}},CH_{t})$
\Else
\State $C_{t}$=\textsc{CheckMerge}$(HS_{t},C_{CH_{t}},CH_{t})$
\State $\left[Q_{t},Y_{Q_{t}}\right]$=\textsc{ActiveQuery}$(HS_{t},C_{t},CH_{t})$
\State $Y_{CH_{t}}$=\textsc{Classify}$(HS_{t},Q_{t},Y_{Q_{t}},CH_{t})$
\State $HS_{t}$=\textsc{ClusteringModel}$(HS_{t},C_{CH_{t}},CH_{t})$
\EndIf
\State \textbf{Return} $HS_{t}$
\EndFor
\end{algorithmic}
\end{algorithm}
\subsection{Concept Drifts and Evolution Detection through Clustering}
To capture the non-stationary property of data streams, we modified the DFPS-clustering algorithm substantially by employing a new cluster merge procedure between the historical clusters and new clusters. Then, we use the new DFPS-clustering method to distinguish novel concepts from drifted concepts. Algorithm \ref{algo:2} summarizes the new cluster merge procedure for the detection of drifted and novel concepts.\\
\indent As shown in Algorithm \ref{algo:2}, the new merge procedure utilizes the density of boundary instances to decide whether a merge should happen between a historical cluster and its neighboring cluster from $CH_{t}$. Then, it generates two types of clusters: (i) \textit{novel clusters} and (ii) \textit{updated clusters}. Clusters that are not merged with historical clusters are considered as \textit{novel clusters} while the merged clusters are defined as \textit{updated clusters}. Based on these two types of clusters, the novel concepts are captured by \textit{novel clusters} and drifted concepts are identified as \textit{updated clusters}. To validate the existence of \textit{novel clusters}, we use the mean and standard deviation of the density values of historical clusters to compute a dynamic density threshold. Let $C^{
no}_{t}$ be a \textit{novel cluster} and $F_{C^{no}_{t}}$ be its density value. The mean and standard deviations of historical clusters are denoted as $\mu(F_{C_{t}})$ and $\sigma(F_{C_{t}})$, respectively. The following Remark is defined for novel cluster detection.
\begin{remark}
If $F_{C^{no}_{t}}\geq |\mu(F_{C_{t}})-\sigma(F_{C_{t}})|$, then $C^{no}_{t}$ is a true novel cluster. \label{remark1}
\end{remark}
\textcolor{black}{Remark \ref{remark1} is derived from the three-sigma principle of Gaussian distribution used in \cite{yan2019novel} and it means a cluster is considered as a valid \textit{novel cluster} if its density value falls inside the one-sigma distance from the average density of historical clusters at time $t$. With Remark \ref{remark1}, a novel cluster detector with a dynamic density threshold is used for validating the existence of \textit{novel clusters}.}
\begin{algorithm}[tb]
\footnotesize
\caption{New cluster merge procedure}
\label{algo:2}
\textbf{Parameters}: $L_{NC}$: a list of paired clusters between a historical cluster and its neighboring clusters in $CH_{t}$, $X_{B}$: a set of boundary samples between each pair of clusters in $L_{NC}$ 
\begin{algorithmic}[1]
\Procedure{CheckMerge}{$HS_{t}$, $C_{CH_{t}}$, $CH_{t}$}
\State Identify the paired neighboring historical clusters in $HS_{t}$ for $C_{CH_{t}}$ to obtain $L_{NC}$
\State Extract $X_{B}$ for each pair of neighboring clusters in $L_{NC}$
\State Evaluate the density of $X_{B}$ and check for the density drop for each pair of neighboring clusters
\State Merge each pair of neighboring clusters when there is no density drop in $X_{B}$
\State Mark unmerged clusters as \textit{novel clusters} and merged clusters as \textit{updated clusters}
\State Validate the existence of \textit{novel clusters} using Remark \ref{remark1}
\State \textbf{Return} $C_{t}=\left[novel \text{ } clusters, updated \text{ } clusters\right]$
\EndProcedure
\end{algorithmic}
\end{algorithm}
\subsection{Adaptation of Drifted and Novel Concepts using Active Learning}
Considering time and space constraints, we maintain a summary of the data stream rather than keeping all historical samples from $DS$. The maintained summary $HS_{t}$ consists of two levels of summary on $DS$: (i). macro-level summary $GS_{t}$, and (ii) micro-level summary $LS_{t}$. The definitions of these two levels of summary are expressed as:
\begin{equation}
    GS_{t} = \bigcup^{|C_{t}|}_{i=1} [Z^{i}_{t},F^{i}_{t},R^{i}_{t}],
\end{equation}
\indent and
\begin{equation}
    LS_{t} = \bigcup^{|C_{t}|}_{i=1} \{\bigcup^{|SC^{i}_{t}|}_{j=1}[SC^{i,j}_{t},y^{i,j}_{t}]\}.
\end{equation}
Where $F^{i}_{t}$ and $R^{i}_{t}$ denote the density value and radius of the $i^{th}$ cluster at time $t$, respectively. $y^{i,j}_{t}$ refers to the class label of $SC^{i,j}_{t}$ and all samples from $SC^{i,j}_{t}$ share the same label. 
$LS_{t}$ is used to explore the sub-cluster structure of each cluster when classes are highly overlapped. Specifically, we split each cluster into a set of sub-clusters such that each sub-cluster only has a unique class label. 
Instead of computing the mean vector for each sub-cluster, we consider only the sample with the highest density value in each sub-cluster as the sub-cluster center and use it for label propagation.

These two levels of summary are continuously updated to adapt to the change of $DS$ using an active learning procedure. Since two types of clusters can be obtained from the clustering analysis, a hybrid active learning strategy of informative-based and representative-based sampling is introduced to reduce the labeling costs. The adaptation procedure of these two levels of summary is provided in Algorithm \ref{algo:3}. In Algorithm \ref{algo:3}, for \textit{novel clusters}, the representative-based query \cite{gu2019active} is performed by sampling from the centers of clusters. 
On the other hand, we conduct the informative-based query \cite{gu2019active} in \textit{updated clusters} through a distance-based strategy. Unlike the entropy-based sampling \cite{lughofer2016recognizing} strategy, samples that are relatively far from the \textit{updated clusters} are selected as informative samples for label querying. Let $Q_{t}$ be the set of queried samples and $Y_{Q_{t}}$ be the label set for $Q_{t}$. After the active label query, the label propagation procedure begins to predict the label of the remaining samples in $CH_{t}$ using $Y_{Q_{t}}$ and $HS_{t}$. Finally, the predicted labels are used to update the $LS_{t}$ with a two-step procedure. First, we update the centers of sub-clusters within \textit{updated clusters} with new samples that have higher density values. Second, we create a set of new sub-clusters for  each \textit{novel cluster} to capture the characteristics of novel concepts. 
\begin{algorithm}[tb]
\footnotesize
\caption{Adaptation of drifted and novel concepts}
\label{algo:3}
\textbf{Parameters:} $X_{no}$: a set of samples from \textit{novel clusters}; $X_{up}$: a set of samples from \textit{updated clusters}; $Q_{I}$: a set of queried samples using the Informative-based sampling; $Q_{R}$: a set of queried samples using Representative-based sampling; $Y_{CH_{t}}$: the label set for $CH_{t}$.
\begin{algorithmic}[1]
\Procedure{ClusteringModel}{$HS_{t}$, $C_{t}$, $CH_{t}$}
\State $\left[Q_{t},Y_{Q_{t}}\right]$=\textsc{ActiveQuery}($C_{t}$, $CH_{t}$)
\State $Y_{CH_{t}}$=\textsc{Classify}$(HS_{t}, Q_{t}, Y_{Q_{t}}, CH_{t})$
\State Update the $GS_{t}$ according to $C_{t}$
\State Update the $LS_{t}$ using $Y_{CH_{t}}$
\State \textbf{return} $HS_{t}=[GS_{t},LS_{t}]$
\EndProcedure
\Procedure{ActiveQuery}{$C_{t},CH_{t}$}
\State Extract \textit{novel clusters} and \textit{updated clusters} from $C_{t}$
\State Identify samples that are close to the \textit{novel clusters} as $X_{no}$
\State Identify samples that are close to \textit{updated clusters} as $X_{up}$
\State Representative-based sampling for $X_{no}$ to obtain $Q_{R}$
\State Informative-based sampling for $X_{up}$ to obtain $Q_{I}$
\State $Q_{t} = (Q_{I}\cup Q_{R})$
\State Query labels from human experts to obtain $Y_{Q_{t}}$
\State \textbf{return} $\left[Q_{t},Y_{Q_{t}}\right]$
\EndProcedure
\end{algorithmic}
\end{algorithm}
\subsection{Classification through Label Propagation}
To classify an incoming data chunk, we employ an effective label propagation procedure based on $HS_{t}$ and $Q_{t}$. First, a set of prototypes with label information are obtained from $HS_{t}$ and $Q_{t}$. Then, the KNN-based classification procedure is employed to propagate the labels of the prototypes to samples in $CH_{t}$. Here, we set the value of $k$ to five and the
classification procedure is presented in Algorithm \ref{algo:4}.
\section{Complexity Analysis} \label{sec:4}
Here, we discussed the time and memory complexity of the CDSC-AL framework using the following parameters:
\begin{itemize}
    \item $n$: the number of samples from an incoming data chunk
    \item $m$: the number of dimensions of an incoming instance
    \item $|SC_{t}|$: the total number of sub-clusters at time $t$
    \item $|C_{t}|$: the total number of clusters at time $t$
    \item $k$: the number of nearest neighbors
    \item $|Q_{t}|$ the number of queried samples at time $t$
\end{itemize}
\begin{algorithm}[tb]
\footnotesize
\caption{Classification through label propagation}
\label{algo:4}
\textbf{Parameters:} $Y_{CH_{t}}$: the label set for $CH_{t}$; $Sub_{r}$: a set of representatives from sub-clusters; $P_{t}$: a set of prototypes with labels.
\begin{algorithmic}[1]
\Procedure{Classify}{$HS_{t},Q_{t},Y_{Q_{t}}, CH_{t}$}
\State Extract sub-cluster centers and its labels from $HS_{t}$ as $Sub_{r}$
\State $P_{t}=Sub_{r}\cup [Q_{t},Y_{Q_{t}}]$
\State Propagate labels from the prototype set to samples in $CH_{t}$ using KNN rule and obtain the predicted label set $Y_{CH_{t}}$ 
\State \textbf{return} $Y_{CH_{t}}$
\EndProcedure
\end{algorithmic}
\end{algorithm}
\paragraph{Time Complexity.}The density evaluation of the new DFPS-clustering procedure requires $O(n^2m)$ distance calculations. To update $HS_{t}$, there will be $O(nm|C_{t}|+nm|SC_{t}|) \leq O(2nm|SC_{t}|)$ distance calculations where $|C_{t}|\leq |SC_{t}|$. The classification of the incoming data chunk will require $O(knm(|SC_{t}|+|Q_{t}|))$ distance calculations. Thus, the total time complexity of a single data chunk in terms of distance calculations is expressed as: $O(n^2m+2nm|SC_{t}|+knm(|SC_{t}|+|Q_{t}|))$.
\paragraph{Memory Complexity.}For $LS_{t}$, the memory space complexity is expressed as $O(|SC_{t}|(m+1)) \approx O(m|SC_{t}|)$. In terms of $GS_{t}$, it takes $O(|C_{t}|(m+2))\approx O(m|C_{t}|)$ memory space. The total memory complexity will be $O(m(|SC_{t}|+|C_{t}|))$.
\section{Experimental Studies and Discussions} \label{sec:5}
In this section, experiments are conducted on the CDSC-AL framework using nine benchmark datasets, and comparison studies with the state-of-the-art methods are presented.
\subsection{Experimental Setup}
\begin{table}[tb]
\footnotesize
\begin{center}
\begin{tabular}{lcccc} %
\hline
\centering
Datasets  & Sample & Dimensions & Classes & Overlap\\
\hline
\centering
Syn-1 & 18900 & 2 & 9 & False \\
\centering
Syn-2 & 11400 & 2 & 10 & True \\
\centering 
Sea & 60000  & 2  & 3 & False  \\
\centering
KDD99  & 494021 & 34 & 5 & False \\  
\centering
Forest & 581012  & 11  &  7 & True  \\ 
\centering
GasSensor & 13910 & 128 & 6 & True \\ 
\centering
Shuttle & 58000  & 9  & 7 & False \\ 
\centering
MNIST & 70000 & 784 & 10 & False \\
\centering
CIFAR10 & 60000 & 3072 & 10 & False \\
\hline
\end{tabular}
\caption{Dataset descriptions.\label{table1}}
\end{center}
\end{table}
\paragraph{Datasets and Streaming Settings.}Nine multi-class benchmark datasets, including three synthetic datasets and six well-known real datasets from \cite{Dua:2017}, are used in the experiments for performance evaluation. Table \ref{table1} summarizes these datasets in terms of sample size, dimensionality, number of classes, and class overlap. According to Table \ref{table1}, the Forest, Syn-2, CIFAR10, and GasSensor datasets have highly overlapped classes. Following the setting in \cite{aggarwal2006framework,din2020online}, we partition each benchmark dataset into a number of data chunks with a fixed size of $1000$ samples and pass them sequentially to each algorithm in the experiment. In Syn-1, Syn-2, Sea, and Shuttle datasets, data chunks are arranged in an order to simulate abrupt concept drifts. For remaining datasets, we arrange data chunks in an order that generates the gradual concept drifts. 
\begin{table}
\centering
\footnotesize
\begin{tabular}{ll|ccc}
\hline
\multirow{1}{*}{Dataset} & \multirow{1}{*}{Metric} & LNP & OReSSL & CDSC-AL  \\
\hline
    \multirow{2}{*}{Syn-1} & \textit{BA} & $0.8869(3)$ & $0.9307(2)$ & $\mathbf{0.9459(1)}$ \\
    & $F_{mac}$ & $0.7939(3)$ & $0.9318(2)$ & $\mathbf{0.9490(1)}$\\
    \hline
    \multirow{2}{*}{Syn-2} & \textit{BA} & $0.8338(3)$ & $\mathbf{0.8481(1)}$ & $0.8459(2)$ \\
    & $F_{mac}$ & $0.6375(3)$ & $0.7899(2)$ & $\mathbf{0.8149(1)}$ \\ 
    \hline
    \multirow{2}{*}{Sea} & \textit{BA} & $0.5262(3)$ & $0.8206(2)$ & $\mathbf{0.9691(1)}$ \\
    & $F_{mac}$ & $0.6019(3)$ & $0.8275(2)$ & $\mathbf{0.9729(1)}$ \\
    \hline
    \multirow{2}{*}{KDD99} & \textit{BA} & $0.5362(3)$ & $0.6831(2)$ & $\mathbf{0.8364(1)}$ \\
    & $F_{mac}$ & $0.5311(3)$ & $0.7076(2)$ & $\mathbf{0.7921(1)}$ \\
    \hline
    \multirow{2}{*}{Forest} & \textit{BA} & $0.5181(3)$ & $0.7153(2)$ & $\mathbf{0.8465(1)}$ \\
    & $F_{mac}$ & $0.5258(3)$ & $0.7114(2)$ & $\mathbf{0.8230(1)}$ \\
    \hline
    \multirow{2}{*}{GasSensor} & \textit{BA} & $0.6479(3)$ & $0.8841(2)$ & $\mathbf{0.8916(1)}$ \\
    & $F_{mac}$ & $0.6611(3)$ & $0.8674(2)$ & $\mathbf{0.8995(1)}$ \\
    \hline
    \multirow{2}{*}{Shuttle} & \textit{BA} & $0.4172(3)$ & $0.4709(2)$ & $\mathbf{0.4744(1)}$ \\
    & $F_{mac}$ & $0.4119(3)$ & $\mathbf{0.4862(1)}$ & $0.4789(2)$ \\
    \hline
    \multirow{2}{*}{MNIST} & \textit{BA} & $0.7682(3)$ & $0.8806(2)$ & $\mathbf{0.9669(1)}$ \\
    & $F_{mac}$ & $0.7725(3)$ & $0.8828(2)$ & $\mathbf{0.9676(1)}$ \\
    \hline
    \multirow{2}{*}{CIFAR10} & \textit{BA} & $0.4158(3)$ & $0.6421(2)$ & $\mathbf{0.7857(1)}$ \\
    & $F_{mac}$ & $0.4195(3)$ & $0.6344(2)$ & $\mathbf{0.7869(1)}$ \\
    \hline
    \hline
    \multirow{2}{*}{Avg. ranks} & \textit{BA} & $3.00$  & $1.78$ & $\mathbf{1.22}$ \\
    & $F_{mac}$ & $3.00$ & $1.89$ & $\mathbf{1.11}$ \\
    \hline
\end{tabular}
\caption{Performance comparison with semi-supervised methods. (Relative rank of each algorithm is shown within parentheses.)\label{table_res1}}
\end{table}
\paragraph{Compared Methods.}We compared the CDSC-AL framework with the state-of-the-art methods considering two aspects: (i) comparison with semi-supervised approaches, and (ii) comparison with supervised approaches. We selected two existing semi-supervised methods, namely OReSSL \cite{din2020online} and LNP \cite{wang2007label}, and results are summarized in Table \ref{table_res1}. Four supervised methods, including Leverage Bagging (LB) \cite{bifet2010leveraging}, OZA Bag ADWIN (OBA) \cite{bifet2009new}, Adaptive Hoeffding Tree (AHT) \cite{bifet2009adaptive}, and SAMkNN \cite{losing2018tackling}, are used for the second comparison study. The results are presented in Table \ref{table_res2}. The MATLAB code for semi-supervised methods is released by the authors and all codes for supervised methods can be found on the Massive Online Analysis (MOA) framework \cite{DBLP:journals/jmlr/BifetHKP10}. The python code of the CDSC-AL framework is available at the link\footnote{https://github.com/XuyangAbert/CDSC-AL}. All experiments are conducted on an Intel Xeon (R) machine with 64GB RAM operating on Microsoft Windows 10.
\paragraph{Parameter Setting.}For the semi-supervised methods and CDSC-AL, the portion of labeled data of each incoming data chunk is set as $10\%$. For supervised methods, the \textit{labels of all samples} from an incoming data chunk are provided to update the classifier after classification while CDSC-AL \textit{utilized only $10\%$ labeled data.} 
\paragraph{Evaluation Metrics.}Due to the imbalanced class distributions in benchmark datasets, we use the \textit{balanced classification accuracy} ($BA$) \cite{brodersen2010balanced} and the macro-average of \textit{F-score} ($F_{mac}$) \cite{kelleher2020fundamentals} as performance evaluation metrics. We recorded these two metrics over the entire data stream classification and reported the average values for performance evaluation. The best results are highlighted in bold-face. The Friedman and Nemenyi post-hoc tests \cite{demvsar2006statistical} are employed to statistically analyze the experimental results with a significance level of $0.05$. 
\begin{table*}
\centering
\footnotesize
\begin{tabular}{ll|ccccc}
\hline
\multirow{1}{*}{Dataset} & \multirow{1}{*}{Metric} & LB & OBA & AHT & SAMkNN & CDSC-AL \\
\hline
    \multirow{2}{*}{Syn-1} & \textit{BA} & $0.7910(2)$ & $0.6640(3)$ & $0.6354(4)$ & $0.6247(5)$ & $\mathbf{0.9459(1)}$\\
    & $F_{mac}$ & $0.7965(2)$ & $0.6675(3)$ & $0.6513(4)$ & $0.6313(5)$ & $\mathbf{0.9490(1)}$\\
    \hline
    \multirow{2}{*}{Syn-2} & \textit{BA} & $0.7124(2)$ & $0.7204(3)$ & $0.6926(4)$ & $0.6784(5)$ & $\mathbf{0.8459(1)}$ \\
    & $F_{mac}$ & $0.7218(2)$ & $0.7219(3)$ & $0.6977(4)$ & $0.6864(5)$ & $\mathbf{0.8149(1)}$ \\
    \hline
    \multirow{2}{*}{Sea} & \textit{BA} & $0.8204(2)$ & $0.7498(3)$ & $0.7493(4)$ & $0.7205(5)$& $\mathbf{0.9691(1)}$  \\
    & $F_{mac}$ & $0.8227(2)$ & $0.7501(4)$ & $0.7505(3)$ &  $0.7345(5)$ & $\mathbf{0.9729(1)}$ \\
    \hline
    \multirow{2}{*}{KDD99} & \textit{BA} & $0.7585(4)$ & $0.7812(3)$ & $\mathbf{0.8541(1)}$ & $0.7495(5)$ & $0.8364(2)$  \\
    & $F_{mac}$ & $0.7564(4)$ & $0.7798(3)$ & $\mathbf{0.8012(1)}$ & $0.7682(5)$ & $0.7921(2)$  \\
    \hline
    \multirow{2}{*}{Forest} & \textit{BA} & $\mathbf{0.8888(1)}$ & $0.8707(2)$ & $0.8612(3)$ & $0.8545(4)$ & $0.8465(5)$  \\
    & $F_{mac}$ & $\mathbf{0.8901(1)}$ & $0.8709(2)$ & $0.8688(3)$ & $0.8588(4)$ & $0.8230(5)$  \\
    \hline
    \multirow{2}{*}{GasSensor} & \textit{BA} & $0.7185(2)$ & $0.6345(4)$ & $0.6111(3)$ & $0.6357(5)$ & $\mathbf{0.8916(1)}$ \\
    & $F_{mac}$ & $0.7199(2)$ & $0.6361(4)$ & $0.6188(3)$ & $0.6412(5)$ & $\mathbf{0.8995(1)}$ \\
    \hline
    \multirow{2}{*}{Shuttle} & \textit{BA} & $\mathbf{0.4789(1)}$ & $0.4477(4)$ & $0.4508(3)$ & $0.4424(5)$ &  $0.4744(2)$  \\
    & $F_{mac}$ & $\mathbf{0.5187(1)}$ & $0.5112(2)$ & $0.4978(3)$ & $0.4894(4)$  & $0.4789(5)$ \\
    \hline
    \multirow{2}{*}{MNIST} & \textit{BA} & $0.8909(2)$ & $0.8498(4)$ & $0.8393(5)$ & $0.8549(3)$ & $\mathbf{0.9669(1)}$  \\
    & $F_{mac}$ & $0.8946(2)$ & $0.8501(4)$ & $0.8412(5)$ & $0.8596(3)$ & $\mathbf{0.9676(1)}$  \\
    \hline
    \multirow{2}{*}{CIFAR10} & \textit{BA} & $0.7199(3)$ & $0.6208(5)$ & $0.7366(2)$ & $0.6218(4)$ & $\mathbf{0.7857(1)}$ \\
    & $F_{mac}$ & $0.7208(2)$ & $0.6325(4)$ & $0.7381(3)$ & $0.6295(5)$ & $\mathbf{0.7869(1)}$ \\
    \hline
    \multirow{2}{*}{Avg. ranks} & \textit{BA} & $2.00$  & $2.86$ & $3.57$ & $4.57$ & $\mathbf{1.86}$ \\
    & $F_{mac}$ & $1.75$ & $2.63$ & $3.00$ & $4.00$ & $\mathbf{2.00}$ \\
    \hline
\end{tabular} 
\caption{Performance comparison with supervised methods. (Relative rank of each algorithm is shown within parentheses.)\label{table_res2}}
\end{table*}
\subsection{Results and Discussions}
\begin{figure}
\centering
\includegraphics[width=55mm,height=18mm]{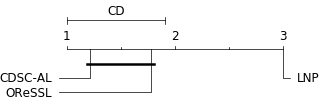}
\caption{Comparison of CDSC-AL against semi-supervised methods with the Nemenyi test with $\alpha=0.05$ using $BA$.}
\label{fig: cd diagrams 1a}
\end{figure}
\begin{figure}
\centering
\includegraphics[width=55mm,height=18mm]{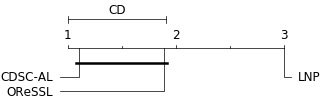}
\caption{Comparison of CDSC-AL against semi-supervised methods with the Nemenyi test with $\alpha=0.05$ using $F_{mac}$.}
\label{fig: cd diagrams 1b}
\end{figure}
\begin{figure}
\centering
\includegraphics[width=60mm,height=22mm]{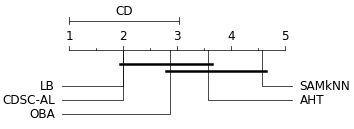}
\caption{Comparison of CDSC-AL against supervised methods with the Nemenyi test with $\alpha=0.05$ using $BA$.}
\label{fig: cd diagrams 2a}
\end{figure}
\begin{figure}
\centering
\includegraphics[width=60mm,height=22mm]{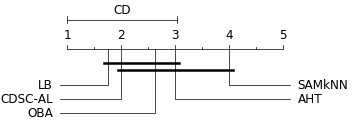}
\caption{Comparison of CDSC-AL against supervised methods with the Nemenyi test with $\alpha=0.05$ using $F_{mac}$.}
\label{fig: cd diagrams 2b}
\end{figure}

\paragraph{Comparison with Semi-supervised Methods.}To compare the CDSC-AL method with the OReSSL and LNP methods, we repeated each experiment ten times and the average results are presented in Table \ref{table_res1}. From Table \ref{table_res1}, the \textit{balanced classification accuracy} shows that the CDSC-AL method outperforms the other two methods on most data streams with the lowest rank of $1.11$. \textcolor{black}{In terms of $F_{mac}$, CDSC-AL also provides better performance on most data streams. For data streams with abrupt concept drifts, CDSC-AL still achieves slightly better or comparable performance. From the Nemenyi post-hoc test, Figures \ref{fig: cd diagrams 1a} and \ref{fig: cd diagrams 1b} reveal that there is a statistically significant difference between CDSC-AL and LNP in terms of $BA$ and $F_{mac}$.} \textcolor{black}{Although CDSC-AL shows statistically comparable performance with the OReSSL method, it does not require any parameter optimization or an initial set of labeled data.}
\paragraph{Comparison with Supervised Methods.}Table \ref{table_res2} presents the results of CDSC-AL and the four supervised methods. Using only $10\%$ of the labels, \textcolor{black}{Table \ref{table_res2} demonstrates that the CDCS-AL method achieves the best performance on six of the benchmark data streams including Syn-1, Syn-2, Sea, GasSensor, MNIST, and CIFAR10. In Figures \ref{fig: cd diagrams 2a} and \ref{fig: cd diagrams 2b}, CDCS-AL shows statistically comparable performance to the LB, OBA, and AHT methods on the remaining three data streams from the Nemenyi test. Also, CDSC-AL has statistically better performance than the SAMkNN approach. For data streams with abrupt concept drifts, CDSC-AL presents slightly better or comparable performance relative to supervised approaches. In summary, the comparison study with supervised methods reveals that CDSDF-AL always provides statistically better or comparable performance than the supervised methods using only a small proportion of labeled data.}
\section{Conclusions and Future Work} \label{sec:6}
We presented a clustering-based data stream classification framework through active learning (CDSC-AL) to handle non-stationary data streams. We developed a new cluster merge procedure as a part of the new DFPS-clustering procedure and used the new DFPS-clustering procedure to capture the drifted concepts and novel concepts. Two levels of cluster summaries were maintained to effectively classify the incoming data samples in the presence of overlapped classes, and an active learning technique was introduced to learn the change of data distributions. The primary advantages of the CDSC-AL method can be summarized from several points of view: (i) no dependency on initial label set, (ii) effective detection for drifted concepts and novel concepts with dynamic threshold, and (iii) high-quality classification performance in the presence of overlapped classes with a small proportion of labeled data. The comparison studies with the semi-supervised and supervised methods justify the efficacy of CDSC-AL methods on both synthetic and real data.\\
\indent In the future, we will investigate solutions to reduce the time complexity of CDSC-AL and implement the CDSC-AL framework in some real-world applications such as text/image stream classifications.
\section*{Acknowledgements}
This work is supported by the Air Force Research Laboratory and the OSD under agreement number FA8750-15-2-0116. Also, this work is partially funded by the NASA University Leadership Initiative (ULI), the National Science Foundation, and the OSD RTL under grants number 80NSSC20M0161, 2000320, and W911NF-20-2-0261 respectively. The authors would like to thank them for their support.
\small
\bibliographystyle{named}
\bibliography{ijcai21}

\end{document}